\documentclass[10pt,twocolumn,letterpaper]{article}

\usepackage[pagenumbers]{cvpr} 
\usepackage{algorithmic}
\usepackage{algorithm}
\usepackage{caption}
\usepackage{float}
\usepackage{amsmath}
\DeclareMathOperator*{\argmax}{arg\,max}  

\newcommand{\expec}{\mathbb{E}}
\newcommand{\encoder}{\mathcal{E}}

\newcommand{\zt}[1]{z_{#1}}

\newcommand{\model}{\epsilon_\theta}
\newcommand{\conditioner}{\tau_\theta}

\usepackage[dvipsnames]{xcolor}

\definecolor{cvprblue}{rgb}{0.21,0.49,0.74}
\usepackage[pagebackref,breaklinks,colorlinks,citecolor=cvprblue]{hyperref}
\DeclareSymbolFont{extraup}{U}{zavm}{m}{n}
\DeclareMathSymbol{\varheart}{\mathalpha}{extraup}{86}
\DeclareMathSymbol{\vardiamond}{\mathalpha}{extraup}{87}
\def\paperID{*****} 
\def\confName{CVPR}
\def\confYear{2024}
\title{iDesigner: A High-Resolution and Complex-Prompt Following Text-to-Image Diffusion Model for Interior Design}

\author{
    \textbf{Ruyi Gan}$^{\varheart\spadesuit}$ \qquad
    \textbf{Xiaojun Wu}$^{\varheart}$ \qquad
    \textbf{Junyu Lu}$^{\varheart}$ \qquad
    \textbf{Yuanhe Tian}$^{\spadesuit}$ \qquad
    \\
    \textbf{Dixiang Zhang}$^{\varheart\vardiamond}$ \qquad
    \textbf{Ziwei Wu}$^{\varheart}$ \qquad
    \textbf{Renliang Sun}$^{\varheart}$ \qquad
    \textbf{Chang Liu}$^{\spadesuit}$ \qquad
    \\
    \textbf{Jiaxing Zhang}$^{\varheart}$\footnotemark[1] \qquad
    \textbf{Pingjian Zhang}$^{\vardiamond}$ \qquad
    \textbf{Yan Song}$^{\spadesuit}$\footnotemark[1] \qquad
    \\
    \\
    $^{\varheart}$International Digital Economy Academy \quad
    $^{\vardiamond}$South China University of Technology \quad \\
    $^{\spadesuit}$University of Science and Technology of China \quad \\
    {\tt\small \{ganruyi, wuxiaojun, lujunyu, zhangdixiang, zhangjiaxing, wuziwei, sunrenliang\}@idea.edu.cn } \\
    {\tt\small yhtian@uw.edu} \quad {\tt\small lc980413@mail.ustc.edu.cn}  \quad {\tt\small pjzhang@scut.edu.cn} \quad {\tt\small clksong@gmail.com}
}

\begin{document}
\maketitle

{
  \renewcommand{\thefootnote}%
    {\fnsymbol{footnote}}
  \footnotetext[1]{Corresponding Authors.}
}

\begin{figure*}[t]
    \centering
    \includegraphics[width=0.90\linewidth, trim=0 10 0 0]{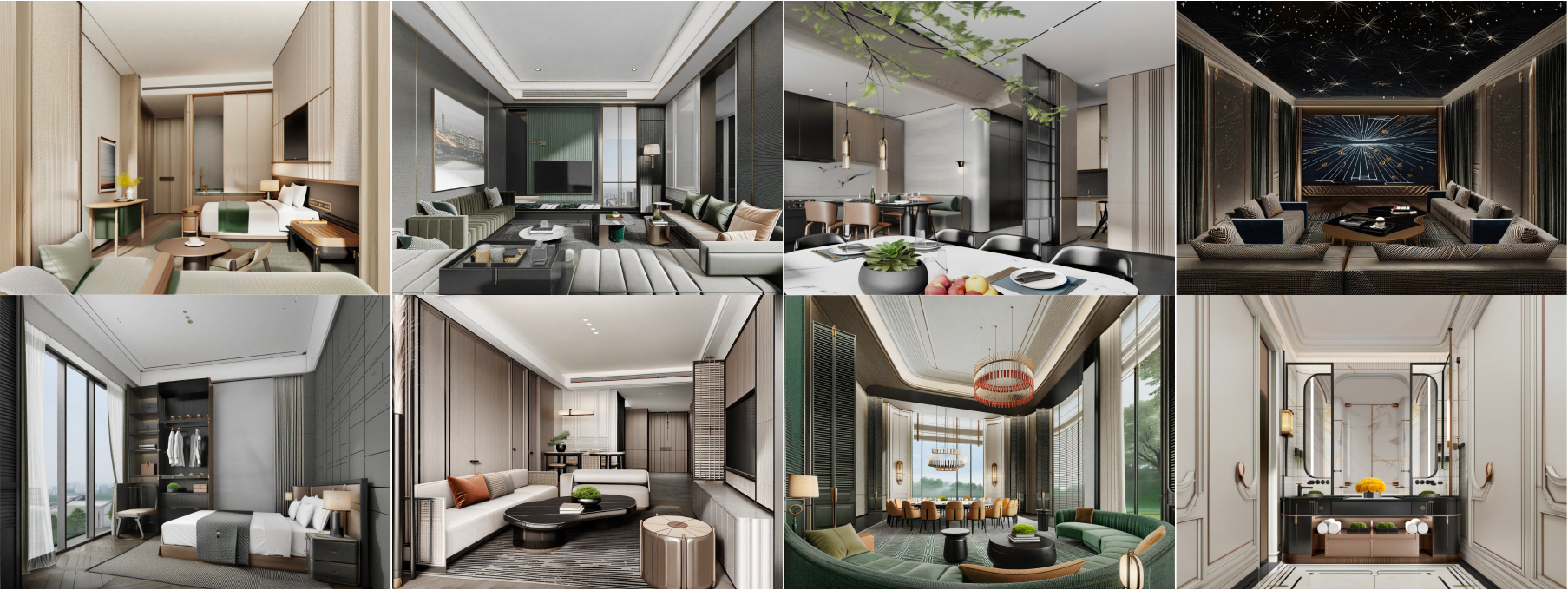}
    \caption{
    Example images generated by the proposed iDesigner model.
    }
    \label{fig: intro example}
    \vskip -1em
\end{figure*}
\begin{abstract}
With the open-sourcing of text-to-image models (T2I) such as stable diffusion (SD) and stable diffusion XL (SD-XL), there is an influx of models fine-tuned in specific domains based on the open-source SD model, such as in anime, character portraits, etc. 
However, there are few specialized models in certain domains, such as interior design, which is attributed to the complex textual descriptions and detailed visual elements inherent in design, alongside the necessity for adaptable resolution.
Therefore, text-to-image models for interior design are required to have outstanding prompt-following capabilities, as well as iterative collaboration with design professionals to achieve the desired outcome.
In this paper, we collect and optimize text-image data in the design field and continue training in both English and Chinese on the basis of the open-source CLIP model. 
We also proposed a fine-tuning strategy with curriculum learning and reinforcement learning from CLIP feedback to enhance the prompt-following capabilities of our approach so as to improve the quality of image generation.
%
The experimental results on the collected dataset demonstrate the effectiveness of the proposed approach, which achieves impressive results and outperforms strong baselines.\footnote{We will release the code and model soon.}

%


\end{abstract}    
\section{Introduction}
\label{sec:intro}






Interior design synthesizes form, function, and aesthetics within physical spaces, requiring meticulous attention to detail and a deep understanding of cultural and contextual elements \cite{chen2023interior}. While text-to-image (T2I) models have made strides in general media generation \cite{cao2022survey, Rombach_2022_CVPR_sd, podell2023sdxl}, they have not been intricately tailored to meet the unique demands of interior design. This disconnect stems from the models' limited ability to process the rich and often complex vocabulary specific to interior design lexicon. As such, the resultant images frequently lack the sophistication and exactitude that professional design work commands. 
%
In this paper, we propose iDesigner, which is designed to generate visually rich and contextually accurate interior design images directly from descriptive text prompts.

iDesigner is crafted to specifically address the unique and complex demands of interior design prompts, a challenge that general-purpose models often fail to meet, resulting in images that fall short of capturing the true essence of the designer's vision. Our model backbone is satble diffsuion XL (SD-XL) which is the most popular T2I open-sourced model in the world \cite{Rombach_2022_CVPR_sd, podell2023sdxl}.
The methodological backbone of iDesigner is its innovative use of curriculum learning \cite{bengio2009curriculum}, a pedagogically inspired approach that scales the complexity of tasks in a graduated manner. It begins with the generation of basic, low-resolution images to ground the model in the foundational aspects of design aesthetics and function. As the model's proficiency escalates, the curriculum progresses to sophisticated, high-resolution image creation, paying meticulous attention to fine details that define the quality and accuracy of professional interior design imagery. Complementing this, iDesigner incorporates a specialized captioner module \cite{chatgpt,james2022dalle3} designed to parse and optimize complex textual prompts. This enables the model to produce images with a higher fidelity to the designer's intent. The integration of the Reinforced Learning from CLIP Feedback (RLCF) method \cite{black2023training, rl_diffusion_survey, janner2022planning, venkatraman2023reasoning} further refines the model's ability to follow prompts, establishing a reinforcing loop between textual instructions and image content that enhances the precision and relevance of the generated images.
Figure \ref{fig: intro example} presents images generated by iDesigner, which demonstrate the effectiveness of iDesigner. More images generated by iDesigner are presented in here \footnote{The images generated by iDesigner are presented in Appendix \ref{appendix_1}.}.
%
The contributions of this paper are summarized as follows:
\begin{itemize}
\item A novel text-to-image model, iDesigner, tailored for the interior design domain and made available to the community for collaborative enhancement and research.
\item A strategic application of prompt engineering and Large Language Models (LLMs) to produce more detailed and vivid captions, which markedly enhances the quality of the generated images.  
\item A pioneering integration of a curriculum learning framework with a diffusion model that progressively sharpens the model's generative capabilities, ensuring superior image quality. While diffusion models are known for their proficiency in generating detailed images, our curriculum-based approach elevates this to a new level. 
\item A RLCF method that reinforces the model's prompt adherence, fine-tuning image generation in accordance with detailed textual instructions.
\end{itemize}

\section{The Approach}
\label{sec:method}
The optimization process of our iDesigner model primarily encompasses several key components. Initially, we annotate interior design renderings across multiple dimensions and employ GPT-3.5 \cite{chatgpt} for rewriting, facilitating the training of our text-to-image model. Subsequently, we conduct a two-stage training process on the CLIP model, yielding a foundation model that is both universally applicable and enhanced for the design domain. Furthermore, this foundational model replaces the textual component of the SD-XL model. We then progressively increase image resolution in a curriculum learning approach during fine-tuning, thereby augmenting the effectiveness of the text-to-image generation. Finally, our trained CLIP model is utilized as a feedback mechanism in reinforcement learning training, significantly improving the model's ability to follow instructions within the interior design domain.
\subsection{Dataset}
\paragraph{Data in Intreior Design Domain}
Due to the current scarcity of high-quality datasets in the field of interior design in the academic community, we established a collaborative partnership with an internationally acclaimed interior design company\footnote{https://www.raritag.com/}, leveraging their extensive historical design records. Our ambitious annotation effort involved the participation of over 1,000 seasoned interior design professionals who meticulously labeled the data. This painstaking process resulted in the curation of a high-quality dataset comprising 3,600 high-quality image-text pairs. To ensure robust model training and evaluation, we carefully divided the dataset into training and testing sets, maintaining a balanced ratio of 9:1. This rigorous data collection and partitioning strategy laid the foundation for our research, enabling us to explore the intricate relationships between text and images in the domain of interior design.

\paragraph{Data Recaptioning}

Our dataset in interior design composed of high-quantity pairings $(X, Y)$, where $X$ represent an image and $Y$ is a text composed of multiple parallel labels that describes the image. In the field of interior design, $Y$ generally comes from the discretized tags marked by designers or derived from web-crawled source, focusing on the simple description of materials, styles and colors. Worse, the web-crawled resource oftentimes contains irrelevant tags, which cannot accurately describe the images. Since the discretized tags inevitably overlook the spatial layout and local details of interior design, and irrelevant tags may mislead the cognition of models, we theorize that such shortcomings can be addressed using synthetically generated captions.
For this purpose, We first collect and structure various interior design datasets. Then, we refer to DALL-E 3 \cite{james2022dalle3} and formulate a set of well-designed system prompts to invoke the GPT3.5 interface for rewriting the alt text into descriptive synthetic captions.


\subsection{CLIP Training}

The vision-language foundation models such as CLIP~\cite{radford2021clip} are crucial component for aligning images and text representations, which can capture the correlation between cross-modal features. Since the open-source CLIP model cannot meet the requirements of bilingual adaptation and multi-element cognition in interior design, We initialize from the pre-trained English-only CLIP and continue training in two stages. In the first stage, we collect a large-scale, web-crawled set of bilingual image-text pairs, including Laion \cite{schuhmann2021laion}, Wukong \cite{gu2022wukong}, and make effort to clean the data. We take the contrastive loss as training objective and utilize a distributed memory-efficient CLIP training approach to reduce the memory consumption~\cite{chen2023discoclip}. In the second stage, we continue to train our CLIP using high-quality common and interior design image-text pairs preprocessed by the image captioner. In interior design domain, the same image may match with multiple distinctly different texts, observing the image from different perspectives and details. 

\subsection{iDesigner Training}
This section primarily discusses the core modules involved in the training of the iDesigner model. Firstly, we introduce the fundamental process of text-to-image generation and describe how we replaced the textual encoder part of the open-source SD-XL with our previously trained CLIP model. Secondly, we detail our approach to curriculum learning and the reinforcement learning method based on CLIP feedback.
\begin{figure*}[t]
    \centering
    \includegraphics[width=0.90\linewidth]{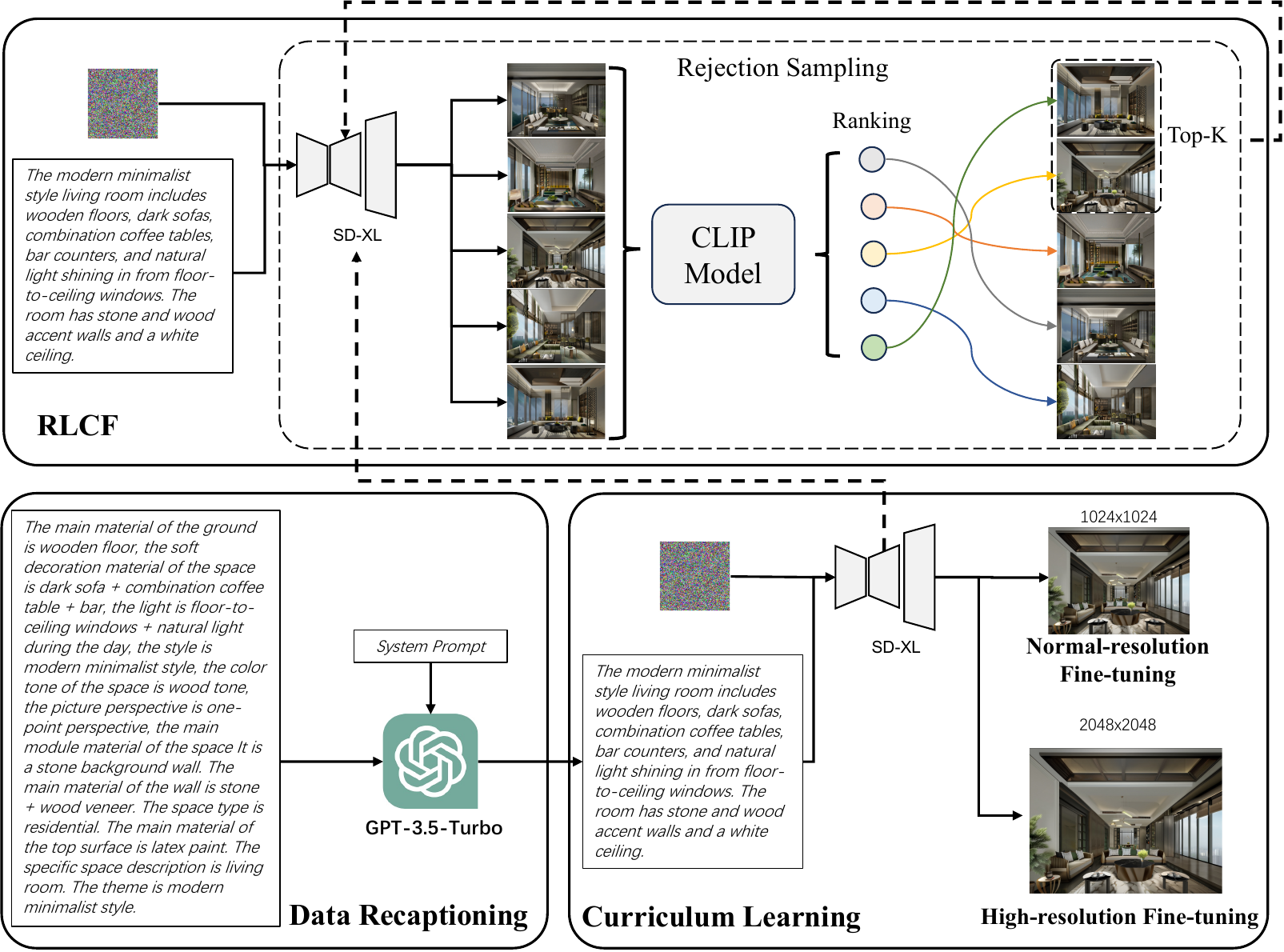}
    \caption{
    An illustration of the overall training process of the iDesigner, which includes data recaptioning, curriculum learning, and reinforcement learning from clip feedback (RLCF)
    }
    \label{fig: structure}
\end{figure*}

\subsubsection{Text-to-image Generation Process}
In the realm of text-to-image generation, particularly with diffusion models, the methodology can be broadly categorized into two primary phases:

\paragraph{Text Encoding}

Traditional models often employ the CLIP text encoder for feature extraction from textual descriptions.
For Chinese-specific applications, the CLIP text encoder is replaced with a dedicated Chinese encoder. This adaptation ensures better alignment with Chinese linguistic structures and semantics.

\paragraph{Text-to-image Generation}

Once textual features are extracted, they are incorporated into the latent diffusion process.
The diffusion process in the latent space offers computational efficiency, reducing both processing time and memory requirements.
The final phase involves training on specific text-image datasets, with the aim of refining the model's capability to generate images that closely match the input textual descriptions.

\subsubsection{Curriculum Learning}


In the context of iDesigner, the nuanced domain of interior design requires a model that is not only proficient in generating images but also exceptional in rendering high-resolution images where even the smallest design elements are crisply defined. To this end, iDesigner employs a curriculum learning (CL) approach, inspired by the pioneering concept introduced by \cite{bengio2009curriculum}, which mirrors the human learning process of progressing from simple to complex levels of understanding. Let us denote the function $\mathcal{G}_\theta$ as the mapping from textual descriptions to image outputs parameterized by \( \theta \), the set of parameters of our model. In the realm of non-convex optimization problems, curriculum learning has demonstrated a substantial enhancement in performance along with a robust generalization capacity. The strategy underpinning curriculum learning expedites the rate of convergence as well as facilitates the discovery of superior local optima within the context of non-convex optimization landscapes. 
We denote \( (x_i, y_i) \) as the paired association of an image \( x_i \) and its corresponding textual descriptor \( y_i \), which together constitute the \( i \)-th instance within the training corpus. The curriculum comprises two steps, each with a dedicated loss function to optimize. For the initial step (Step 1) at resolution \( 1024 \times 1024 \), we define a loss function \( L_1 \) which focuses on the global structure and basic elements of design space:
\begin{equation}
 L_1(\theta) := \expec_{\encoder(x_{low}), y, \epsilon \sim \mathcal{N}(0, 1), t }\Big[ \Vert \epsilon - \model(z_{t},t, \conditioner(y)) \Vert_{2}^{2}\Big] \, ,
\label{eq:cond_loss}
\end{equation}
where \( x_{low} \) is the downsampled version of \( x_i \) to the lower resolution and \( \text{loss}_{1} \) measures the difference between the generated low-resolution image and the ground truth.  
These models can be interpreted as an equally weighted sequence of denoising latent autoencoders $\model(\zt{t},t);\, t=1\dots T$ where the $\model$ is realized as a time-conditional UNet \cite{ronneberger2015unet}. . 
Since the forward process is fixed, ${t}$ can be efficiently obtained from
$\encoder(x)$ during training and can be decoded to image space with a single pass through vae decoder \cite{kingma2013vae}. 
And $\conditioner$ can be parameterized with transformer text encoder model, both $\conditioner$ and $\model$ are jointly optimized via Eq.~\ref{eq:cond_loss}.
This foundational step allows the model to establish an understanding of the broader aesthetic and functional principles of interior design without being overwhelmed by the intricacies of high-resolution details.


Once the model has demonstrated proficiency in generating coherent and contextually accurate images at this resolution, the curriculum progresses to Step 2, where the resolution is elevated to \( 2048 \times 2048 \) at resolution. We introduce a loss function \( L_2 \) which aims at refining the generated images to capture detailed design elements:
%
\begin{equation}
L_2(\theta) := \expec_{\encoder(x_{high}), y, \epsilon \sim \mathcal{N}(0, 1), t }\Big[ \Vert \epsilon - \model(z_{t},t, \conditioner(y)) \Vert_{2}^{2}\Big] \, ,
\label{eq:cond_loss2}
\end{equation}
where \( L_2 \) quantifies the fidelity of the generated high-resolution image in terms of texture, pattern detail, and local design element accuracy. This higher resolution phase challenges the model to refine its generative capabilities, focusing on the minute details such as textures of fabrics, the play of light on different surfaces, and the precise appearance of small furniture and objects that are pivotal in a realistic interior design rendering. 

The overall curriculum learning process can be described by a compound loss function \( L \) over the course of training epochs \( E \), which is a combination of \( L_1 \) and \( L_2 \) with a weighting function \( \alpha(e) \) that adjusts the contribution of each loss function over time:
\begin{equation}
L(\theta, e) = \alpha(e) \cdot L_1(\theta) + (1 - \alpha(e)) \cdot L_2(\theta)
\label{eq:total_loss}
\end{equation}
Herein, \( \alpha(e) \) is a monotonically decreasing function with respect to epoch \( e \), where \( \alpha(0) = 1 \) at the beginning of training and gradually decreases to \( 0 \) as training proceeds. This effectively shifts the training focus from the global structure to the intricate details as the model's capacity increases.

The model parameters \( \theta \) are updated iteratively using a stochastic gradient descent method or one of its variants to minimize the loss function \( L(\theta, e) \) over epochs:
%
\begin{equation}
\theta_{e+1} = \theta_e - \eta \cdot \nabla_{\theta}L(\theta_e, e)
\label{eq:backpropagate_loss}
\end{equation}
where \( \eta \) is the learning rate and \( \nabla_{\theta} \) denotes the gradient with respect to the parameters \( \theta \).

This incremental approach is crucial, as our experiments have shown that directly training a model at \( 2048 \times 2048 \) pixels on interior design data, when the underlying base model is trained at \( 1024 \times 1024 \), leads to structural imbalances in the generated images. These imbalances manifest as discrepancies in the spatial arrangement of furniture, inconsistencies in texture and pattern detail, and a general loss of image cohesion. The curriculum learning strategy in iDesigner effectively prevents such imbalances by allowing the model to develop a hierarchical understanding of interior design elements. It starts with mastering the global layout and composition before delving into the local details. The use of curriculum learning in iDesigner not only ensures that the generated images maintain structural integrity at higher resolutions but also greatly improves the model’s generalization capabilities and convergence rates, a testament to the efficacy of curriculum learning strategies in complex interior design domains of text-to-image.

This formalism ensures that iDesigner is guided through a structured learning pathway, handling increasing levels of complexity in a controlled manner that is reflective of the staged learning process in human education. By employing this staged approach, iDesigner avoids the pitfalls of structural imbalances that occur when training directly at higher resolutions. The curriculum learning technique ensures the model progressively acquires a nuanced understanding of interior design elements, from global layout to intricate details, thereby maintaining image integrity and improving the model’s generalization capabilities and convergence rates, demonstrating the power of curriculum learning strategies in the complex domain of interior design in text-to-image synthesis. The overall training algorithm of iDesigner is summarized in Algorithm \ref{alg:iDesigner_training}.


\begin{algorithm}[t]
\caption{Curriculum Learning for iDesigner}
\label{alg:iDesigner_training}
\begin{algorithmic}[1]
\STATE Initialize iDesigner $\mathcal{G}_\theta$ with noise predictor $\model$, text encoder $\conditioner$, latent encoder $\encoder$, and dataset $\mathcal{D}$
\FOR{epoch in range(${n_{epochs}}$)}
    \STATE Sample a batch of data $\mathcal{B}$ from $\mathcal{D}$
    \FOR{$(x_i, y_i)\in\mathcal{B}$}
        \STATE Generate text embeddings $\conditioner(y_i)$ 
        \STATE Generate latent embeddings $\zt{i} = \encoder(x_{i})$ 
        \STATE $t \sim \text{Uniform}(\{1, \ldots, T\})$
        \STATE $\epsilon \sim \mathcal{N}(0, I)$
        \STATE Obtain $\zt{i_t}$ by adding noise ${\epsilon}$ to $\zt{i}$  
        \STATE Feed $(\zt{i_t},\conditioner(y_i))$ to the $\model$ to generate noise predictions ${\epsilon}_{pred}$
        \STATE Compute $\mathcal{L}$ over the batch $\mathcal{B}$ between ${\epsilon}$ and ${\epsilon}_{pred}$ using Eq. (\ref{eq:total_loss})
    \STATE Update $\theta$ by backpropagating $\mathcal{L}$ using Eq. (\ref{eq:backpropagate_loss})
    \ENDFOR
\ENDFOR
\STATE \textbf{return} the fine-tuned model $\mathcal{G}_\theta$.
\end{algorithmic}
\end{algorithm}

\subsubsection{RLCF: Reinforcement Learning From CLIP Feedback}
As the prompts in the field of interior design consist of various tags, even after the aforementioned Data recaption and two-stage SFT in the design field, the images generated by iDesigner still show an increased relevance with the text, but there are still mismatches between the details of the text and the images. Therefore, we hope to introduce reinforcement learning, which has achieved tremendous success on LLMs, to enhance the effect. Unlike ChatGPT \cite{chatgpt} and LLaMA2 \cite{touvron2023llama2}, which use human feedback, we directly employ CLIP, fine-tuned in the design field, for model feedback, and then iterate using the method of reject sampling. In each iteration, we score the images generated by iDesigner and the original text with CLIP. The higher the score, the greater the relevance between the image and text. We select the TopK images based on the CLIP scores, and along with the original image and text, we form a new set of (K+1) image-text pairs for fine-tuning iDesigner. 
We now delineate the RLCF (Reinforced Learning from CLIP Feedback) algorithm, which is structured into three distinct steps for each stage $t+1$:

\textbf{Step 1: Data collection.} A batch of prompts $\mathcal{D}_t=\{y_1^t,\cdots,y_b^t\}$ is sampled from the text domain, and for each prompt $y_i^t \in \mathcal{D}_t$, a set of images $x_1,\ldots,x_K$ are generated by the image synthesis model iDesigner $\mathcal{G}$.

\textbf{Step 2: Data ranking.} Using the reward model CLIP, we compute a set of rewards $\{r(y,x_1),\cdots,r(y,x_K)\}$ for each prompt $y \in \mathcal{D}_t$. Subsequently, we select the image with the highest reward: $x := \argmax_{x_j \in \{x_1,\cdots, x_K\}} r(y, x_j)$, and repeat this for all $b$ prompts to form a subset $\mathcal{B}$ of size $b$.

\textbf{Step 3: Model fine-tuning.} The iDesigner model $\mathcal{G}_\theta$ is fine-tuned on the subset $\mathcal{B}$, and thereafter, the next stage of the learning process commences.

This iterative process continues until the reward, as determined by the CLIP model, converges. The RLCF algorithm boasts minimal hyperparameter tuning and is straightforward to implement. It capitalizes on the best-of-$K$ policy, where the model iteratively learns to produce image samples that are increasingly aligned with the highest rewards as gauged by CLIP, thereby refining the iDesigner model's generation capabilities.

\section{Experiment and Result}
\noindent\textbf{Training Settings.}
For our iDesigner model, we employ the pre-trained checkpoint of Stable Diffusion XL (SD-XL) \cite{podell2023sdxl} as our foundational backbone, ensuring a robust starting point for image generation tasks. To optimize resource utilization and expedite the training process, we utilize the BFLOAT16 format, which significantly reduces GPU memory requirements while maintaining training efficiency. Our training regimen adopts a learning rate of 1e-5, which is carefully calibrated through a warmup phase to stabilize the learning dynamics initially. This is followed by a cosine decay schedule to gradually reduce the learning rate, facilitating fine-tuning and convergence to a more precise model state. These settings are critical in achieving the delicate balance between training speed and model performance.

\noindent\textbf{Baselines.}
In our comparative analysis, we consider two strong baselines: DALL-E 3 \cite{james2022dalle3} and SD-XL \cite{podell2023sdxl}. DALL-E 3 is renowned for its innovative text-to-image capabilities, generating high-quality images from textual descriptions. It serves as a benchmark for cutting-edge generative models. SD-XL, on the other hand, is a variant of the Stable Diffusion model known for its extended capabilities in handling complex image synthesis tasks. By comparing iDesigner with these established models, we aim to demonstrate the effectiveness and advancements of our approach, particularly in terms of bilingual image generation and adherence to textual prompts.

\noindent\textbf{Evaluation Protocols.}
Our evaluation framework encompasses both machine and human assessments to provide a comprehensive understanding of the model's performance. Machine evaluation metrics include CLIP performance evaluation with image-to-text retrieval and text-to-image retrieval; CLIP Similarity (CLIP Sim), which measures the semantic alignment between the generated images and text descriptions; Inception Score (IS), assessing the quality and diversity of the images; and Fréchet Inception Distance (FID), evaluating the distance between the distributions of generated and real images. Human evaluation, on the other hand, involves subjective assessments by a group of evaluators. They rate the images based on visual appeal, relevance to the provided prompts, and overall aesthetic quality. This dual approach ensures a well-rounded evaluation, combining objective computational assessments with human perceptual judgments.

\subsection{Machine Evaluation}

The performance of our CLIP model\footnote{The result detail of CLIP  is presented in Appendix \ref{appendix_2}} achieves the best performance on both English and Chinese datasets. On the Flickr \cite{young2014flickr} and MSCOCO datasets \cite{lin2014coco} cross-lingual image-text retrieval tasks, the original CLIP model demonstrates a foundational understanding, with modest retrieval rates that highlight the challenge of transferring learning across languages. In contrast, AltCLIP \cite{chen2022altCLIP} and our-CLIP exhibit remarkable improvements, with our-CLIP attaining the highest recall rates across most metrics. Notably, in the Text $\rightarrow$ Image retrieval task, our-CLIP achieves an impressive 88.1\% and 69.7\% recall at 1 on the Flickr-CN \cite{young2014flickr} and MSCOCO-CN datasets \cite{li2019coco-cn} respectively, indicating a robust alignment between text prompts and visual content. These results underscore the efficacy of tailored modifications to enhance CLIP's cross-lingual performance, emphasizing the potential of specialized models in handling diverse linguistic contexts within multimodal AI applications.

The presented data in Table \ref{table:models_comparison_bilingual} offers a comprehensive overview of the performance of various models across English and Chinese datasets, evaluated using CLIP Similarity (CLIP Sim), Inception Score (IS), and Fréchet Inception Distance (FID) metrics. Notably, the iDesigner\_RLCF model demonstrates superior performance in both linguistic contexts, achieving the highest CLIP Sim, IS, and the lowest FID scores, distinctly outperforming other models like SD-XL and DALL-E 3.

In the English dataset, iDesigner\_RLCF attains a CLIP Sim of 0.145, indicating a more refined alignment between text and image compared to SD-XL and DALL-E 3, which score 0.112 and 0.118, respectively. This enhancement in semantic consistency is further corroborated by its IS score of 4.690, surpassing iDesigner\_1024 and iDesigner\_2048 variants, suggesting a more accurate capture of nuanced image details. The model’s proficiency in generating high-quality images is also reflected in its FID score of 76.832, the lowest among the compared models, indicating closer proximity to real image distributions.

The trend is consistent in the Chinese dataset, where iDesigner\_RLCF again leads with a CLIP Sim of 0.136, an IS of 4.315, and an FID of 78.102. Compared to the English dataset, a slight variation in scores is observed, possibly attributing to the linguistic and cultural differences inherent in the datasets. However, the model’s robustness across languages is evident, marking a significant advancement in bilingual image generation capabilities.

These results collectively underline the effectiveness of the RLCF approach in iDesigner, particularly in enhancing the model’s ability to comprehend and respond to complex textual prompts accurately, thereby generating images that are not only visually appealing but also semantically coherent across diverse linguistic contexts.

\begin{table}[t]
\centering
\small
\begin{tabular}{lccc}
\toprule
Model & CLIP Sim($\uparrow$)  & IS($\uparrow$) & FID($\downarrow$) \\
\cmidrule(r){1-4}
\multicolumn{4}{c}{English Dataset} \\
\midrule
Test Set & 0.205 & 4.838 & 0 \\
SD-XL~\cite{podell2023sdxl} & 0.112 & 3.450 & 95.867 \\
DALL-E 3~\cite{james2022dalle3} & 0.118  & 3.832 & 89.906 \\
iDesigner\_1024 & 0.135  & 4.562 & 79.340 \\
iDesigner\_2048 & 0.137  & 4.559 & 79.262 \\
iDesigner\_RLCF & \textbf{0.145}  & \textbf{4.690} & \textbf{76.832} \\
\midrule
\multicolumn{4}{c}{Chinese Dataset} \\
\cmidrule(r){1-4}
Test Set & 0.181  & 4.838 & 0 \\
SD-XL~\cite{podell2023sdxl} & 0.096 & 3.007 & 95.439 \\
DALL-E 3~\cite{james2022dalle3} & 0.106  & 3.201 & 90.236 \\
iDesigner\_1024 & 0.129  & 4.004 & 80.172 \\
iDesigner\_2048 & 0.126  & 4.309 & 79.816 \\
iDesigner\_RLCF & \textbf{0.136}  & \textbf{4.315} & \textbf{78.102} \\
\bottomrule
\end{tabular}
\caption{\label{table:models_comparison_bilingual}
Comparison of different models based on CLIP Sim and IS and FID across English and Chinese datasets.The best results except the original Test Set are marked in \textbf{bold}.}
\end{table}


\subsection{Human Preference Evaluation}

\begin{figure}
\centering
\includegraphics[width=.48\textwidth, keepaspectratio]{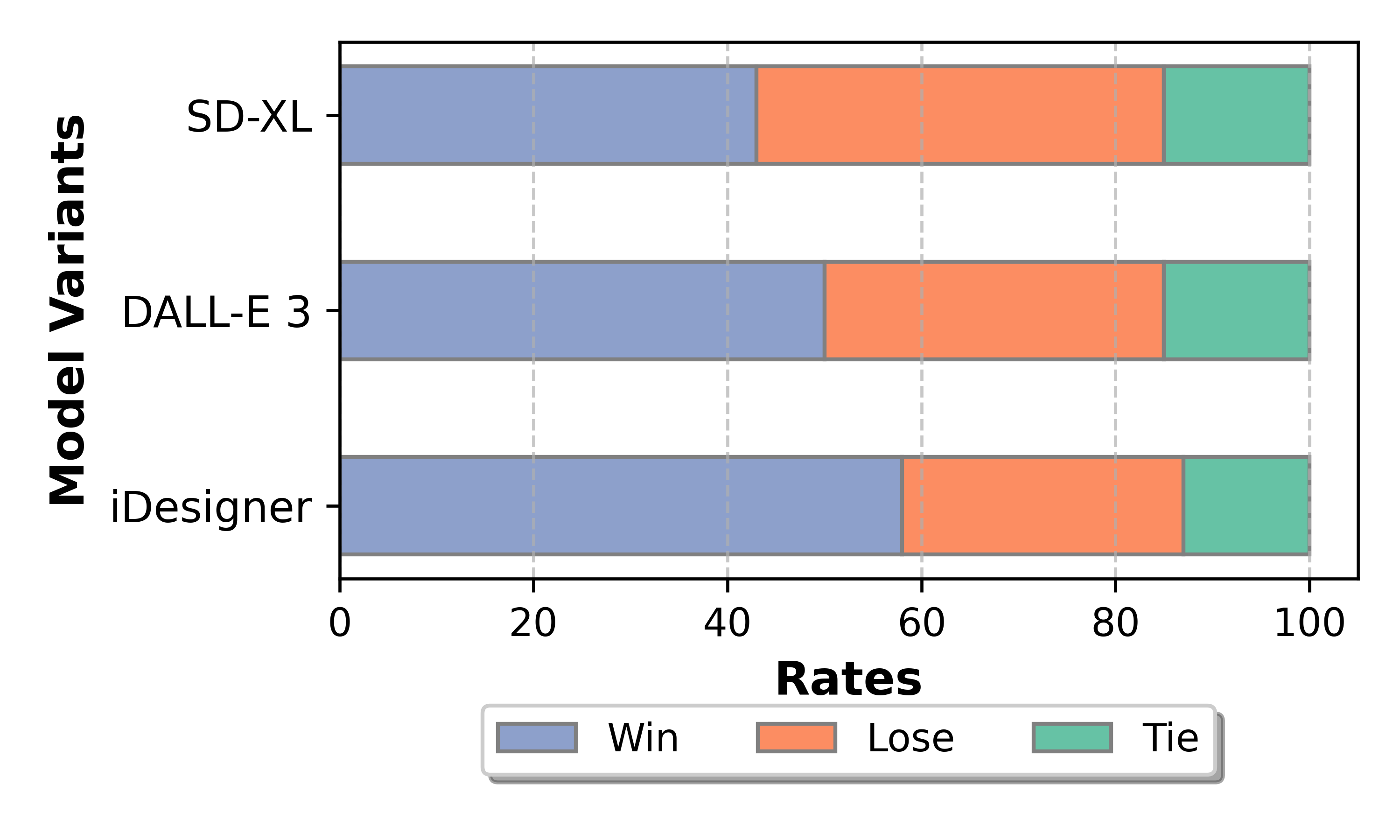}
\caption{Results of the human preference evaluation, illustrating the win/lose/tie rates of our iDesigner method against the other competing models. The iDesigner model demonstrates a clear preference among human evaluators, substantiating its superior image generation capabilities.}
\label{human-result}
\end{figure}

\begin{figure*}[h]
\centering
\includegraphics[width=0.98\textwidth, keepaspectratio]
{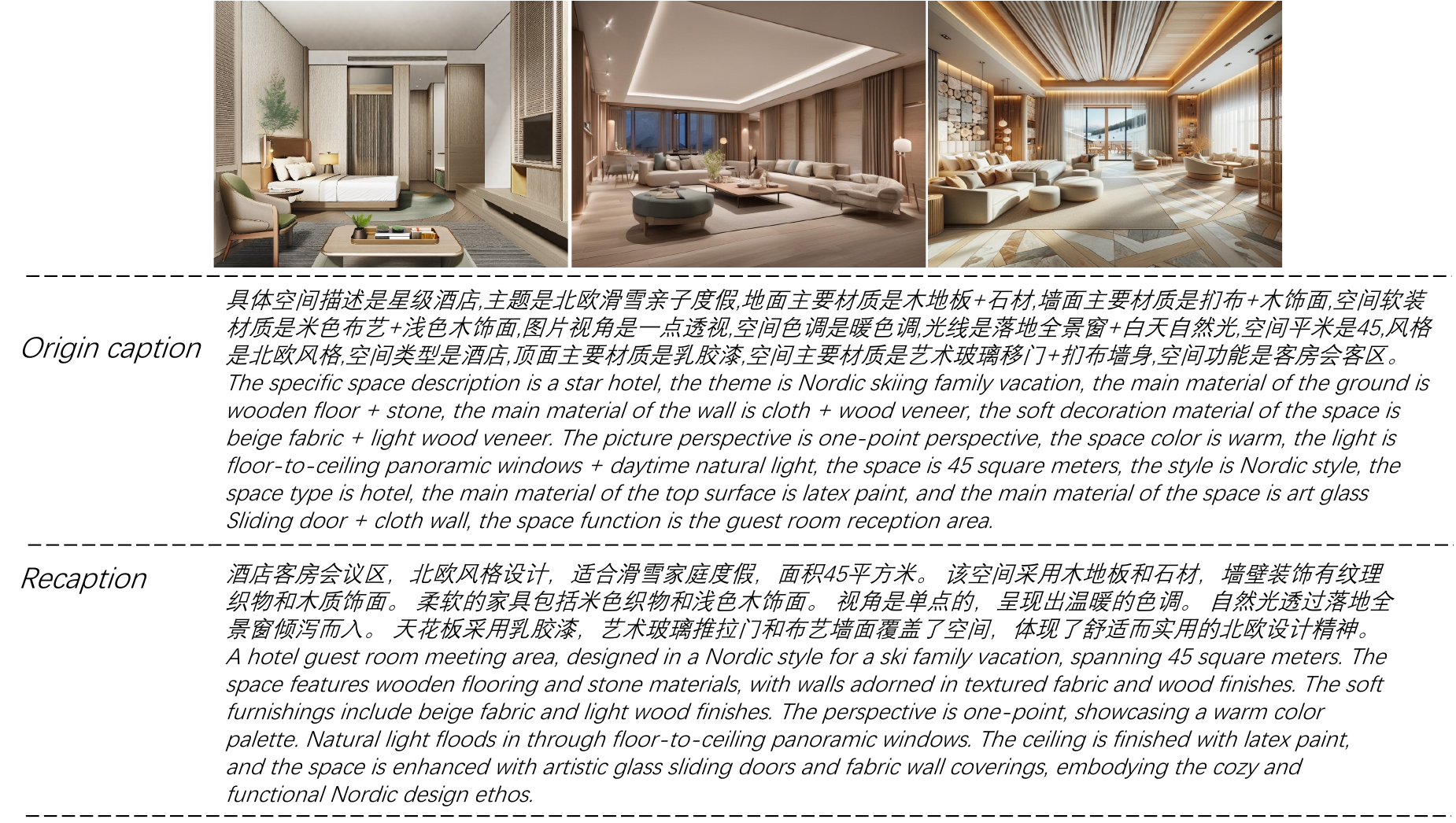}
\caption{The images, from left to right, are generated respectively by iDesigner, SD-XL, and DALL-E 3.}
\label{imgs}
\end{figure*}

In addition to the automated metrics presented in Table \ref{table:models_comparison_bilingual}, we performs a human preference evaluation to directly assess the perceptual quality of the generated images. Participants in the study were asked to compare images generated by iDesigner, DALL-E 3, and SD-XL models in response to a variety of prompts. The participants were blinded to the model origins of the images to prevent any potential bias. The results, as depicted in Figure \ref{human-result}, unequivocally indicate a preference for iDesigner, with the model achieving a win rate of over 58\% against both DALL-E 3 and SD-XL baselines. This substantial margin not only reinforces the quantitative findings from the CLIP scores but also illustrates the qualitative leap in generation fidelity that iDesigner represents. The generated images were frequently cited as more coherent, aesthetically pleasing, and true to the textual descriptions provided.
In Figure \ref{imgs}, we present images generated by the same prompt using iDesigner, SD-XL and DALL-E 3. Compared to other T2I models, our results are more closely aligned with the stylistic essence of human designer output and exhibit superior image quality.

This subjective evaluation underscores the effectiveness of iDesigner in preserving the semantic essence of the original prompts while manifesting images that resonate more strongly with human judges. The ability of iDesigner to maintain high CLIP scores while also securing human preference attests to its advanced capability in generating semantically and visually compelling images. The feedback from human evaluators provides invaluable insights that go beyond numerical scores, highlighting the nuanced improvements that iDesigner brings to the realm of text-to-image synthesis.

\subsection{Ablation Study}


In our ablation study, as shown in Table \ref{tab:ablation_study}, we systematically evaluate the contribution of each component in the iDesigner model. The complete iDesigner model with RLCF achieves the highest scores across all metrics. The removal of RLCF results in a notable decrease in CLIP Score to 0.120, highlighting its significant role in improving text-image semantic alignment. Removing Curriculum Learning impacts both the Inception Score and FID, with a decrease to 4.200 and an increase to 85.000, respectively, indicating its importance in enhancing image quality and diversity. The absence of the captioner has a milder effect, evidenced by a slight decrease in the Inception Score to 4.500 and a marginal increase in FID to 76.900. These results collectively illustrate the synergistic effect of these components in optimizing the performance of the iDesigner model, with each playing a critical role in achieving high-quality, semantically coherent image generation.

\begin{table}[t]
\centering
\small
\begin{tabular}{lccc}
\toprule
Model Variant & CLIP Sim($\uparrow$)  & IS($\uparrow$) & FID($\downarrow$) \\
\midrule
iDesigner & \textbf{0.145} & \textbf{4.690} & \textbf{76.832} \\
iDesigner w/o Cap & 0.142 & 4.500 & 76.900 \\
iDesigner w/o CL & 0.140 & 4.200 & 85.000 \\
iDesigner w/o RLCF & 0.120 & 4.650 & 77.100 \\
\bottomrule
\end{tabular}
\caption{\label{tab:ablation_study}
Ablation study on iDesigner. Cap: Captioner, CL: Curriculum Learing, RLCF: Reinforcement learning From CLIP Feedback.}
\end{table}

Overall, the ablation study underscores the integral role of each component in achieving the final performance of iDesigner. The harmonized interplay between the captioner, curriculum learning, and reject sampling is what endows iDesigner with its remarkable ability to generate images that are not only visually compelling but also deeply resonant with the textual prompts provided by users.

\section{Related Work}

\paragraph{Image Generation and Diffusion Model}The field of text-to-image generation has witnessed significant advancements in recent years. Compared to earlier models such as GAN \cite{goodfellow2014gan, arjovsky2017wgan}, VAE \cite{kingma2013vae}, Flow-based model \cite{rezende2015flow}, and autoregressive models \cite{ramesh2021dalle, ding2021cogview, ding2022cogview2}, this work places a greater emphasis on the advanced diffusion model. With the advancement and maturation of diffusion theory and techniques \cite{vincent2011dsm, ho2020ddpm, song2020ddim, cao2022survey}, the diffusion model has started to become one of the mainstream technologies in the field of image generation. Notable developments include: Dall-E 2 \cite{ramesh2022dalle2} introduces a hierarchical approach to generate images conditioned on textual descriptions using CLIP latents while \cite{james2022dalle3} pointed out that  better captions can improving image generation quality. Imagen \cite{saharia2022imagen} and Deepfloyd-IF \cite{alex2023deep-floyd_IF} presents a diffusion model that generates photorealistic images from textual descriptions, emphasizing deep language understanding. The current most popular diffusion model is the latent diffusion model \cite{Rombach_2022_CVPR_sd}, which includes a series of works such as stable-diffusion-v1-5, stable-diffusion-2-1, and stable-diffusion-xl \cite{podell2023sdxl}. These model primarily extracts textual features using the CLIP text model and then incorporates these textual features into the latent diffusion process. Conducting the diffusion process in the latent space can reduce computational overhead and memory requirements. Moreover, due to significant advancements in reinforcement learning within large language models, there have been attempts \cite{janner2022planning,black2023training,venkatraman2023reasoning,rl_diffusion_survey} to integrate reinforcement learning into extended models. This integration aims to enhance the quality of generation and the degree of control over text. Although diffusion models have been employed in various design fields such as character design, scene design, and architectural design, most applications are based on simple fine-tuning of the stable diffusion model. Particularly, given the multitude of elements and the complexity of scenes characteristic of interior design, there remains a lack of a customized text-to-image model specifically tailored for the interior design domain.

\paragraph{Bilingual Text-to-image model}
Chinese researchers, aiming to better adapt to the text-to-image needs in Chinese scenarios, have proposed numerous works. The mainstream Chinese diffusion image generation models are mostly derived from further training based on stable-diffusion. This typically involves two steps. The first step is to replace the CLIP text encoder with a bilingual encoder or Chinese encoder, followed by pre-training for text-image matching on a Chinese text-image dataset. Representative works in this category include Taiyi-CLIP \cite{zhang2022fengshenbang}, Chinese-CLIP \cite{yang2022chineseclip}, and Alt-CLIP \cite{chen2022altCLIP}. The second step involves replacing the text encoder in stable diffusion and then continuing the training on a Chinese text-image dataset for text-to-image generation. As a result, we obtain the Chinese version of the diffusion image generation model, with representative works like Taiyi-diffusion \cite{zhang2022fengshenbang} and Alt-diffusion \cite{Ye2023AltDiffusionAM}. However, replacing the CLIP text encoder often means that the entire text-to-image model will lose its English capabilities, and the training cost can be relatively high.

\paragraph{Text-image dataset} Whether in text-image matching or text-to-image generation, datasets play a crucial role. Traditional image caption datasets, such as MSCOCO \cite{lin2014coco} and Flickr \cite{young2014flickr} in the English domain and MSCOCO-CN \cite{li2019coco-cn} and Flickr-CN \cite{li2016fickr-cn} in the Chinese domain, are suitable for training but have relatively smaller sizes, typically below one million. As a result, web-crawled datasets like Laion \cite{schuhmann2021laion} (primarily in English) and Wukong \cite{gu2022wukong} (primarily in Chinese) have become more critical sources of data for training diffusion text-to-image models. These datasets have reached scales of 100 million or even 5 billion.



\vspace{0.1cm}

\section{Conclusion}

\vspace{0.1cm}

In this paper, we propose the first Chinese-English bilingual text-to-image model in the interior design field, and optimize the data processing and model training methods to meet the needs of the design field such as data scarcity, complex text descriptions and image elements, and diverse resolution pictures, and finally obtaine a good generation effect in this field. In the future, we will continue to try to optimize the model in the following directions, including continuing to increase the quantity and quality of text-image data, accessing the knowledge of large language models, and using multidimensional feedback reinforcement learning.

{
    \small
    \bibliographystyle{ieeenat_fullname}
    \bibliography{main}

\begin{thebibliography}{38}
\providecommand{\natexlab}[1]{#1}
\providecommand{\url}[1]{\texttt{#1}}
\expandafter\ifx\csname urlstyle\endcsname\relax
  \providecommand{\doi}[1]{doi: #1}\else
  \providecommand{\doi}{doi: \begingroup \urlstyle{rm}\Url}\fi

\bibitem[Alex~Shonenkov(2023)]{alex2023deep-floyd_IF}
Daria Bakshandaeva Christoph Schuhmann Ksenia Ivanova Nadiia~Klokova Alex~Shonenkov, Misha~Konstantinov.
\newblock If: Title of the repository, 2023.

\bibitem[Arjovsky et~al.(2017)Arjovsky, Chintala, and Bottou]{arjovsky2017wgan}
Martin Arjovsky, Soumith Chintala, and L{\'e}on Bottou.
\newblock Wasserstein generative adversarial networks.
\newblock In \emph{International conference on machine learning}, pages 214--223. PMLR, 2017.

\bibitem[Bengio et~al.(2009)Bengio, Louradour, Collobert, and Weston]{bengio2009curriculum}
Yoshua Bengio, J{\'e}r{\^o}me Louradour, Ronan Collobert, and Jason Weston.
\newblock Curriculum learning.
\newblock In \emph{Proceedings of the 26th annual international conference on machine learning}, pages 41--48, 2009.

\bibitem[Black et~al.(2023)Black, Janner, Du, Kostrikov, and Levine]{black2023training}
Kevin Black, Michael Janner, Yilun Du, Ilya Kostrikov, and Sergey Levine.
\newblock Training diffusion models with reinforcement learning.
\newblock \emph{arXiv preprint arXiv:2305.13301}, 2023.

\bibitem[Cao et~al.(2022)Cao, Tan, Gao, Chen, Heng, and Li]{cao2022survey}
Hanqun Cao, Cheng Tan, Zhangyang Gao, Guangyong Chen, Pheng-Ann Heng, and Stan~Z Li.
\newblock A survey on generative diffusion model.
\newblock \emph{arXiv preprint arXiv:2209.02646}, 2022.

\bibitem[Chen et~al.(2023{\natexlab{a}})Chen, Shao, and Hu]{chen2023interior}
Junming Chen, Zichun Shao, and Bin Hu.
\newblock Generating interior design from text: A new diffusion model-based method for efficient creative design.
\newblock \emph{Buildings}, 13\penalty0 (7):\penalty0 1861, 2023{\natexlab{a}}.

\bibitem[Chen et~al.(2023{\natexlab{b}})Chen, Qi, Wang, and Zhang]{chen2023discoclip}
Yihao Chen, Xianbiao Qi, Jianan Wang, and Lei Zhang.
\newblock Disco-clip: A distributed contrastive loss for memory efficient clip training.
\newblock In \emph{Proceedings of the IEEE/CVF Conference on Computer Vision and Pattern Recognition}, pages 22648--22657, 2023{\natexlab{b}}.

\bibitem[Chen et~al.(2022)Chen, Liu, Zhang, Ye, Yang, and Wu]{chen2022altCLIP}
Zhongzhi Chen, Guang Liu, Bo-Wen Zhang, Fulong Ye, Qinghong Yang, and Ledell Wu.
\newblock Altclip: Altering the language encoder in clip for extended language capabilities.
\newblock \emph{arXiv preprint arXiv:2211.06679}, 2022.

\bibitem[Ding et~al.(2021)Ding, Yang, Hong, Zheng, Zhou, Yin, Lin, Zou, Shao, Yang, et~al.]{ding2021cogview}
Ming Ding, Zhuoyi Yang, Wenyi Hong, Wendi Zheng, Chang Zhou, Da Yin, Junyang Lin, Xu Zou, Zhou Shao, Hongxia Yang, et~al.
\newblock Cogview: Mastering text-to-image generation via transformers.
\newblock \emph{Advances in Neural Information Processing Systems}, 34:\penalty0 19822--19835, 2021.

\bibitem[Ding et~al.(2022)Ding, Zheng, Hong, and Tang]{ding2022cogview2}
Ming Ding, Wendi Zheng, Wenyi Hong, and Jie Tang.
\newblock Cogview2: Faster and better text-to-image generation via hierarchical transformers.
\newblock \emph{Advances in Neural Information Processing Systems}, 35:\penalty0 16890--16902, 2022.

\bibitem[Goodfellow et~al.(2014)Goodfellow, Pouget-Abadie, Mirza, Xu, Warde-Farley, Ozair, Courville, and Bengio]{goodfellow2014gan}
Ian Goodfellow, Jean Pouget-Abadie, Mehdi Mirza, Bing Xu, David Warde-Farley, Sherjil Ozair, Aaron Courville, and Yoshua Bengio.
\newblock Generative adversarial nets.
\newblock \emph{Advances in neural information processing systems}, 27, 2014.

\bibitem[Gu et~al.(2022)Gu, Meng, Lu, Hou, Minzhe, Liang, Yao, Huang, Zhang, Jiang, et~al.]{gu2022wukong}
Jiaxi Gu, Xiaojun Meng, Guansong Lu, Lu Hou, Niu Minzhe, Xiaodan Liang, Lewei Yao, Runhui Huang, Wei Zhang, Xin Jiang, et~al.
\newblock Wukong: A 100 million large-scale chinese cross-modal pre-training benchmark.
\newblock \emph{Advances in Neural Information Processing Systems}, 35:\penalty0 26418--26431, 2022.

\bibitem[Ho et~al.(2020)Ho, Jain, and Abbeel]{ho2020ddpm}
Jonathan Ho, Ajay Jain, and Pieter Abbeel.
\newblock Denoising diffusion probabilistic models.
\newblock \emph{Advances in neural information processing systems}, 33:\penalty0 6840--6851, 2020.

\bibitem[James~Betker(2023)]{james2022dalle3}
Li~Jing Tim Brooks Jianfeng Wang Linjie Li Long Ouyang Juntang Zhuang Joyce Lee Yufei Guo Wesam Manassra Prafulla Dhariwal Casey Chu Yunxin Jiao Aditya~Ramesh James~Betker, Gabriel~Goh.
\newblock Improving image generation with better captions.
\newblock \emph{openai cdn.openai.com/papers/dall-e-3.pdf}, 2023.

\bibitem[Janner et~al.(2022)Janner, Du, Tenenbaum, and Levine]{janner2022planning}
Michael Janner, Yilun Du, Joshua~B Tenenbaum, and Sergey Levine.
\newblock Planning with diffusion for flexible behavior synthesis.
\newblock \emph{arXiv preprint arXiv:2205.09991}, 2022.

\bibitem[Kingma and Welling(2013)]{kingma2013vae}
Diederik~P Kingma and Max Welling.
\newblock Auto-encoding variational bayes.
\newblock \emph{arXiv preprint arXiv:1312.6114}, 2013.

\bibitem[Li et~al.(2016)Li, Lan, Dong, and Liu]{li2016fickr-cn}
Xirong Li, Weiyu Lan, Jianfeng Dong, and Hailong Liu.
\newblock Adding chinese captions to images.
\newblock In \emph{Proceedings of the 2016 ACM on international conference on multimedia retrieval}, pages 271--275, 2016.

\bibitem[Li et~al.(2019)Li, Xu, Wang, Lan, Jia, Yang, and Xu]{li2019coco-cn}
Xirong Li, Chaoxi Xu, Xiaoxu Wang, Weiyu Lan, Zhengxiong Jia, Gang Yang, and Jieping Xu.
\newblock Coco-cn for cross-lingual image tagging, captioning, and retrieval.
\newblock \emph{IEEE Transactions on Multimedia}, 21\penalty0 (9):\penalty0 2347--2360, 2019.

\bibitem[Lin et~al.(2014)Lin, Maire, Belongie, Hays, Perona, Ramanan, Doll{\'a}r, and Zitnick]{lin2014coco}
Tsung-Yi Lin, Michael Maire, Serge Belongie, James Hays, Pietro Perona, Deva Ramanan, Piotr Doll{\'a}r, and C~Lawrence Zitnick.
\newblock Microsoft coco: Common objects in context.
\newblock In \emph{Computer Vision--ECCV 2014: 13th European Conference, Zurich, Switzerland, September 6-12, 2014, Proceedings, Part V 13}, pages 740--755. Springer, 2014.

\bibitem[OpenAI(2022)]{chatgpt}
OpenAI.
\newblock Introducing {C}hat{G}{P}{T}, 2022.

\bibitem[Podell et~al.(2023)Podell, English, Lacey, Blattmann, Dockhorn, M{\"u}ller, Penna, and Rombach]{podell2023sdxl}
Dustin Podell, Zion English, Kyle Lacey, Andreas Blattmann, Tim Dockhorn, Jonas M{\"u}ller, Joe Penna, and Robin Rombach.
\newblock Sdxl: Improving latent diffusion models for high-resolution image synthesis.
\newblock \emph{arXiv preprint arXiv:2307.01952}, 2023.

\bibitem[Radford et~al.(2021)Radford, Kim, Hallacy, Ramesh, Goh, Agarwal, Sastry, Askell, Mishkin, Clark, et~al.]{radford2021clip}
Alec Radford, Jong~Wook Kim, Chris Hallacy, Aditya Ramesh, Gabriel Goh, Sandhini Agarwal, Girish Sastry, Amanda Askell, Pamela Mishkin, Jack Clark, et~al.
\newblock Learning transferable visual models from natural language supervision.
\newblock In \emph{International conference on machine learning}, pages 8748--8763. PMLR, 2021.

\bibitem[Ramesh et~al.(2021)Ramesh, Pavlov, Goh, Gray, Voss, Radford, Chen, and Sutskever]{ramesh2021dalle}
Aditya Ramesh, Mikhail Pavlov, Gabriel Goh, Scott Gray, Chelsea Voss, Alec Radford, Mark Chen, and Ilya Sutskever.
\newblock Zero-shot text-to-image generation.
\newblock In \emph{International Conference on Machine Learning}, pages 8821--8831. PMLR, 2021.

\bibitem[Ramesh et~al.(2022)Ramesh, Dhariwal, Nichol, Chu, and Chen]{ramesh2022dalle2}
Aditya Ramesh, Prafulla Dhariwal, Alex Nichol, Casey Chu, and Mark Chen.
\newblock Hierarchical text-conditional image generation with clip latents.
\newblock \emph{arXiv preprint arXiv:2204.06125}, 1\penalty0 (2):\penalty0 3, 2022.

\bibitem[Rezende and Mohamed(2015)]{rezende2015flow}
Danilo Rezende and Shakir Mohamed.
\newblock Variational inference with normalizing flows.
\newblock In \emph{International conference on machine learning}, pages 1530--1538. PMLR, 2015.

\bibitem[Rombach et~al.(2022)Rombach, Blattmann, Lorenz, Esser, and Ommer]{Rombach_2022_CVPR_sd}
Robin Rombach, Andreas Blattmann, Dominik Lorenz, Patrick Esser, and Bj\"orn Ommer.
\newblock High-resolution image synthesis with latent diffusion models.
\newblock In \emph{Proceedings of the IEEE/CVF Conference on Computer Vision and Pattern Recognition (CVPR)}, pages 10684--10695, 2022.

\bibitem[Ronneberger et~al.(2015)Ronneberger, Fischer, and Brox]{ronneberger2015unet}
Olaf Ronneberger, Philipp Fischer, and Thomas Brox.
\newblock U-net: Convolutional networks for biomedical image segmentation.
\newblock In \emph{Medical Image Computing and Computer-Assisted Intervention--MICCAI 2015: 18th International Conference, Munich, Germany, October 5-9, 2015, Proceedings, Part III 18}, pages 234--241. Springer, 2015.

\bibitem[Saharia et~al.(2022)Saharia, Chan, Saxena, Li, Whang, Denton, Ghasemipour, Gontijo~Lopes, Karagol~Ayan, Salimans, et~al.]{saharia2022imagen}
Chitwan Saharia, William Chan, Saurabh Saxena, Lala Li, Jay Whang, Emily~L Denton, Kamyar Ghasemipour, Raphael Gontijo~Lopes, Burcu Karagol~Ayan, Tim Salimans, et~al.
\newblock Photorealistic text-to-image diffusion models with deep language understanding.
\newblock \emph{Advances in Neural Information Processing Systems}, 35:\penalty0 36479--36494, 2022.

\bibitem[Schuhmann et~al.(2021)Schuhmann, Vencu, Beaumont, Kaczmarczyk, Mullis, Katta, Coombes, Jitsev, and Komatsuzaki]{schuhmann2021laion}
Christoph Schuhmann, Richard Vencu, Romain Beaumont, Robert Kaczmarczyk, Clayton Mullis, Aarush Katta, Theo Coombes, Jenia Jitsev, and Aran Komatsuzaki.
\newblock Laion-400m: Open dataset of clip-filtered 400 million image-text pairs.
\newblock \emph{arXiv preprint arXiv:2111.02114}, 2021.

\bibitem[Song et~al.(2020)Song, Meng, and Ermon]{song2020ddim}
Jiaming Song, Chenlin Meng, and Stefano Ermon.
\newblock Denoising diffusion implicit models.
\newblock \emph{arXiv preprint arXiv:2010.02502}, 2020.

\bibitem[Touvron et~al.(2023)Touvron, Martin, Stone, Albert, Almahairi, Babaei, Bashlykov, Batra, Bhargava, Bhosale, et~al.]{touvron2023llama2}
Hugo Touvron, Louis Martin, Kevin Stone, Peter Albert, Amjad Almahairi, Yasmine Babaei, Nikolay Bashlykov, Soumya Batra, Prajjwal Bhargava, Shruti Bhosale, et~al.
\newblock Llama 2: Open foundation and fine-tuned chat models.
\newblock \emph{arXiv preprint arXiv:2307.09288}, 2023.

\bibitem[Venkatraman et~al.(2023)Venkatraman, Khaitan, Akella, Dolan, Schneider, and Berseth]{venkatraman2023reasoning}
Siddarth Venkatraman, Shivesh Khaitan, Ravi~Tej Akella, John Dolan, Jeff Schneider, and Glen Berseth.
\newblock Reasoning with latent diffusion in offline reinforcement learning.
\newblock \emph{arXiv preprint arXiv:2309.06599}, 2023.

\bibitem[Vincent(2011)]{vincent2011dsm}
Pascal Vincent.
\newblock A connection between score matching and denoising autoencoders.
\newblock \emph{Neural computation}, 23\penalty0 (7):\penalty0 1661--1674, 2011.

\bibitem[Yang et~al.(2022)Yang, Pan, Lin, Men, Zhang, Zhou, and Zhou]{yang2022chineseclip}
An Yang, Junshu Pan, Junyang Lin, Rui Men, Yichang Zhang, Jingren Zhou, and Chang Zhou.
\newblock Chinese clip: Contrastive vision-language pretraining in chinese.
\newblock \emph{arXiv preprint arXiv:2211.01335}, 2022.

\bibitem[Ye et~al.(2023)Ye, Liu, Wu, and Wu]{Ye2023AltDiffusionAM}
Fulong Ye, Guangyi Liu, Xinya Wu, and Ledell~Yu Wu.
\newblock Altdiffusion: A multilingual text-to-image diffusion model.
\newblock \emph{ArXiv}, abs/2308.09991, 2023.

\bibitem[Young et~al.(2014)Young, Lai, Hodosh, and Hockenmaier]{young2014flickr}
Peter Young, Alice Lai, Micah Hodosh, and Julia Hockenmaier.
\newblock From image descriptions to visual denotations: New similarity metrics for semantic inference over event descriptions.
\newblock \emph{Transactions of the Association for Computational Linguistics}, 2:\penalty0 67--78, 2014.

\bibitem[Zhang et~al.(2022)Zhang, Gan, Wang, Zhang, Zhang, Yang, Gao, Wu, Dong, He, et~al.]{zhang2022fengshenbang}
Jiaxing Zhang, Ruyi Gan, Junjie Wang, Yuxiang Zhang, Lin Zhang, Ping Yang, Xinyu Gao, Ziwei Wu, Xiaoqun Dong, Junqing He, et~al.
\newblock Fengshenbang 1.0: Being the foundation of chinese cognitive intelligence.
\newblock \emph{arXiv preprint arXiv:2209.02970}, 2022.

\bibitem[Zhu et~al.(2023)Zhu, Zhao, He, Zhong, Zhang, Yu, and Zhang]{rl_diffusion_survey}
Zhengbang Zhu, Hanye Zhao, Haoran He, Yichao Zhong, Shenyu Zhang, Yong Yu, and Weinan Zhang.
\newblock Diffusion models for reinforcement learning: A survey.
\newblock \emph{arXiv preprint arXiv:2311.01223}, 2023.

\end{thebibliography}
}


\clearpage
\appendix

\section{Example Images Generated by iDesigner}
\label{appendix_1}

\begin{figure*}[h]
\centering
\includegraphics[width=0.95\textwidth, keepaspectratio]{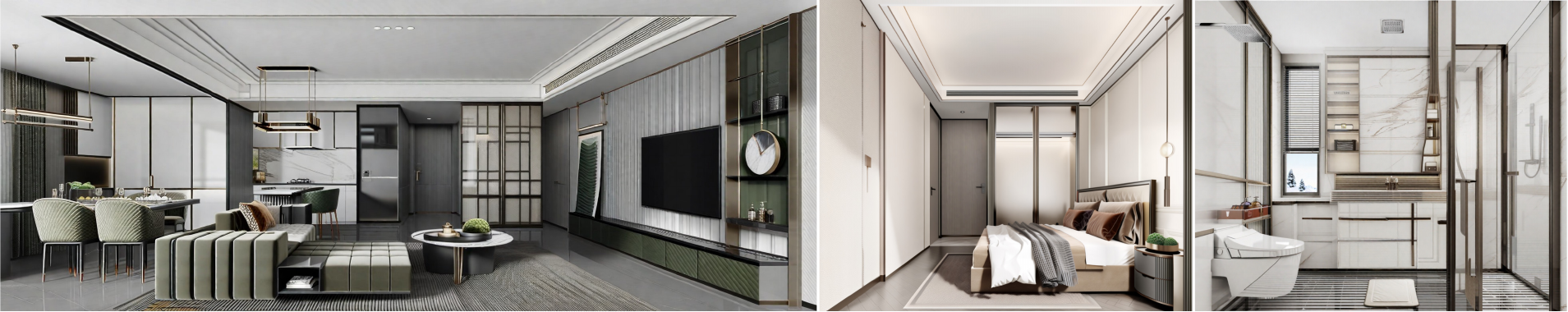}
\caption{Examples of image comparisons with different resolutions which are generated by iDesigner.}
\label{compare_designer}
\end{figure*}

Figure \ref{compare_designer} presents the example images generated by the proposed iDesigner, which illustrate the effective of our approach in generating high-quality images.

\section{CLIP Results in General Dataset}
\label{appendix_2}
Table \ref{tabs:clip_en} and \ref{tabs:clip_zh} presents the performance of our CLIP model on datasets comprising English and Chinese captions respectively. A CLIP model endowed with robust bilingual comprehension capabilities can significantly enhance the ability of our iDesigner to understand user-input prompts and subsequently generate images that accurately conform to the given prompts.

\begin{table*}[h]
\centering
\resizebox{0.95\textwidth}{!}{%
\begin{tabular}{@{}lcccccccccccc@{}}
    \toprule 
    & \multicolumn{6}{c}{Flickr30K} & \multicolumn{6}{c}{MSCOCO} \\
    & \multicolumn{3}{c}{Image $\rightarrow$ Text} & \multicolumn{3}{c}{Text $\rightarrow$ Image} & \multicolumn{3}{c}{Image $\rightarrow$ Text} & \multicolumn{3}{c}{Text $\rightarrow$ Image} \\
    \cmidrule(lr){2-4} \cmidrule(lr){5-7} \cmidrule(lr){8-10} \cmidrule(lr){11-13}
    Model & R@1 & R@5 & R@10 & R@1 & R@5 & R@10 & R@1 & R@5 & R@10 & R@1 & R@5 & R@10 \\
    \midrule
    CLIP \cite{radford2021clip}  & 85.1 & 97.3 & 99.2 &  65.0 & 87.1 & 92.2 &  56.4 & 79.5 & 86.5 &  36.5 & 61.1 & 71.1 \\
    AltCLIP \cite{chen2022altCLIP} & 86.0 & 98.0 & 99.1 &  72.5 & 91.6 & 95.4 &  58.6 & 80.6 & 87.8 &  42.9 & 68.0 & 77.4 \\
    our-CLIP                     & \textbf{88.4} & \textbf{98.8} & \textbf{99.9} &  \textbf{75.7} & \textbf{93.8} & \textbf{96.9} &  \textbf{61.2} & \textbf{84.8} & \textbf{90.3} &  \textbf{49.2} & \textbf{70.3} & \textbf{79.6}  \\
    \bottomrule
\end{tabular}
}
\caption{Zero-shot image-text retrieval results on Flickr30K~\cite{young2014flickr} and MSCOCO~\cite{lin2014coco} datasets.}
\label{tabs:clip_en}
\end{table*}

\begin{table*}{h}
\centering
\resizebox{0.95\textwidth}{!}{%
\begin{tabular}{@{}lcccccccccccc@{}}
    \toprule 
    & \multicolumn{6}{c}{Flickr30K-CN} & \multicolumn{6}{c}{MSCOCO-CN} \\
    & \multicolumn{3}{c}{Image $\rightarrow$ Text} & \multicolumn{3}{c}{Text $\rightarrow$ Image} & \multicolumn{3}{c}{Image $\rightarrow$ Text} & \multicolumn{3}{c}{Text $\rightarrow$ Image} \\
    \cmidrule(lr){2-4} \cmidrule(lr){5-7} \cmidrule(lr){8-10} \cmidrule(lr){11-13}
    Model & R@1 & R@5 & R@10 & R@1 & R@5 & R@10 & R@1 & R@5 & R@10 & R@1 & R@5 & R@10 \\
    \midrule
    CLIP~\cite{radford2021clip}  & 2.3  & 8.1  & 12.6 & 0    & 2.4  & 4.0  & 0.6  & 4.1  & 7.1  &  1.8  &  6.7 &  11.9 \\
    AltCLIP~\cite{chen2022altCLIP} & 69.8 & 89.9 & 94.7 & 84.8 & 97.4 & 98.8 & 63.9 & 87.2 & 93.9 & 62.8  & 88.8 &  95.5\\
    our-CLIP                     & \textbf{73.2} & \textbf{90.3} & \textbf{96.5} & \textbf{88.1} & \textbf{98.2} & \textbf{99.1} & \textbf{66.0} & \textbf{91.1} & \textbf{96.6} & \textbf{69.7}  & \textbf{91.3} &  \textbf{96.8}  \\
    \bottomrule
  
\end{tabular}
}
\caption{\label{tabs:clip_zh} Zero-shot image-text retrieval results on Flickr30K-CN~\cite{young2014flickr} and MSCOCO-CN~\cite{li2019coco-cn} datasets.}
\end{table*}

\end{document}